\documentclass{article} 
\usepackage{iclr2026_conference,times}

\usepackage{amsmath,amsfonts,bm}

\def\eqref#1{equation~\ref{#1}}

\def\1{\bm{1}}

\DeclareMathAlphabet{\mathsfit}{\encodingdefault}{\sfdefault}{m}{sl}
\SetMathAlphabet{\mathsfit}{bold}{\encodingdefault}{\sfdefault}{bx}{n}

\usepackage{hyperref}
\usepackage{url}
\usepackage{multirow}
\usepackage{graphicx}
\usepackage{booktabs}
\usepackage{xcolor}
\usepackage{array}
\usepackage[utf8]{inputenc} 
\usepackage{wrapfig}
\usepackage[T1]{fontenc}    
\usepackage{hyperref}       
\usepackage{url}            
\usepackage{booktabs}       
\usepackage{amsfonts}       
\usepackage{nicefrac}       
\usepackage{microtype}      
\usepackage{xcolor}         
\usepackage{amsmath}
\usepackage{amssymb}
\usepackage{mathtools}
\usepackage{amsthm}
\usepackage{graphicx, subcaption}
\usepackage{tablefootnote}
\usepackage[capitalize,noabbrev]{cleveref}
\usepackage{xspace}
\usepackage[table]{xcolor}
\NewDocumentCommand{\update}{ mO{} }{\textcolor{blue}{\textsf{\textbf{\small#1}}}}
\NewDocumentCommand{\yuzhen}{ mO{} }{\textcolor{red}{\textsuperscript{\textit{yuzhen}}\textsf{\textbf{\small[#1]}}}}
\newcommand{\toolName}{IceCache\xspace}
\newcommand{\revise}[1]{{#1}}
\usepackage{algorithm}
\usepackage{algpseudocode}

\title{IceCache: Memory-efficient KV-cache Management for Long-Sequence LLMs}

\author{Yuzhen Mao \\
Simon Fraser University, BC, Canada\\
\texttt{yuzhenm@sfu.ca} \\
\And
\hspace{-30mm}Qitong Wang\\
\hspace{-30mm}Harvard University, Cambridge, USA \\
\hspace{-30mm}\texttt{qitong@seas.harvard.edu} \\
\And
Martin Ester\\
Simon Fraser University, BC, Canada\\
\texttt{ester@sfu.ca}
\And
\hspace{25mm}Ke Li\\
\hspace{25mm}Simon Fraser University, BC, Canada\\
\hspace{25mm}\texttt{keli@sfu.ca}
}

\iclrfinalcopy
\begin{document}

\maketitle

\begin{abstract}
    Key-Value (KV) cache plays a crucial role in accelerating inference in large language models (LLMs) by storing intermediate attention states and avoiding redundant computation during autoregressive generation. However, its memory footprint scales linearly with sequence length, often leading to severe memory bottlenecks on resource-constrained hardware. Prior work has explored offloading KV-cache to the CPU while retaining only a subset on the GPU, but these approaches often rely on imprecise token selection and suffer performance degradation in long-generation tasks such as chain-of-thought reasoning. In this paper, we propose a novel KV-cache management strategy, \toolName, which integrates semantic token clustering with PagedAttention. By organizing semantically related tokens into contiguous memory regions managed by a hierarchical, dynamically updatable data structure, our method enables more efficient token selection and better utilization of memory bandwidth during CPU–GPU transfers. Experimental results on LongBench show that, with a 256-token budget, \toolName maintains 99\% of the original accuracy achieved by the full KV-cache model. Moreover, compared to other offloading-based methods, \toolName attains competitive or even superior latency and accuracy while using only 25\% of the KV-cache token budget, demonstrating its effectiveness in long-sequence scenarios. The code is available on our project website at~\url{https://yuzhenmao.github.io/IceCache/}.

\end{abstract}

\vspace{-2mm}
\section{Introduction} \label{sec:intro}
  \vspace{-2mm}
Key-Value (KV) cache is a critical component in modern large language models (LLMs), which stores the intermediate attention outputs for each token, allowing the model to reuse these computations in subsequent forward passes. This mechanism is particularly important for autoregressive generation, where tokens are produced sequentially. However, a fundamental challenge of KV-cache lies in its memory consumption: as the generated sequence length increases, the cache size grows linearly, often leading to severe memory pressure or out-of-memory errors on devices with limited resources.

\begin{wrapfigure}{r}{0.5\textwidth}
    \centering
    \vspace{-9mm}
    \includegraphics[width=0.5\textwidth]{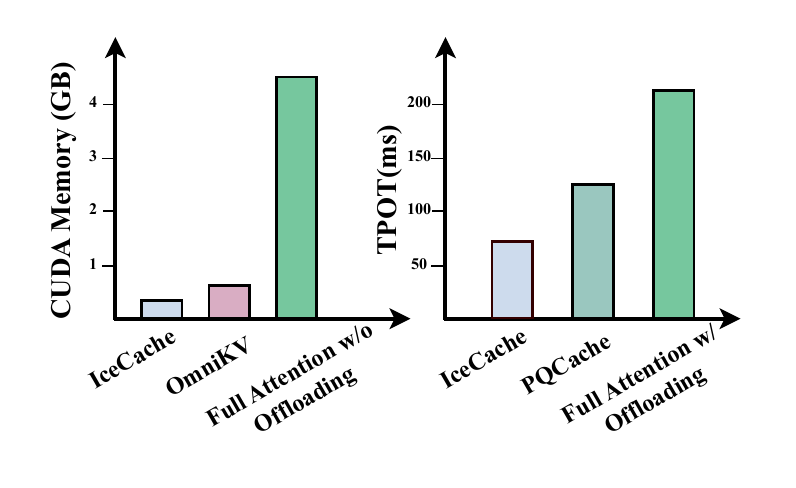}
    \vspace{-10mm}\caption{\toolName has the best trade-off between CUDA memory footprint and time-per-output-token (TPOT) on A100 at a 36k sequence length. Baselines are chosen to represent high-accuracy (Left) and memory-efficient (Right) methods.}
    \vspace{-9mm}
    \label{fig:teaser}
\end{wrapfigure}
Recent studies~\citep{zhang2024h2o, tang2024quest, xiao2023efficient} have shown that, despite the growing size of the KV-cache, only a small subset of tokens contributes disproportionately to generation accuracy. Building on this insight, subsequent work~\citep{chenarkvale,lee2024infinigen,chen2024magicpig} offloads the KV-cache to the CPU while dynamically retaining only the most important entries on the GPU. However, many existing approaches lack precise mechanisms for identifying truly relevant tokens, resulting in low hit rates for the most relevant cache entries. In addition, they often lack effective update mechanisms to manage the KV-cache pool as new tokens are generated during decoding. Therefore, in scenarios involving long-generation tasks, such as long-context summarization, multi-step reasoning, and chain-of-thought (CoT) generation, previous methods experience significant performance degradation~\citep{li2024scbench}.

\begin{figure}[H]
\centering
\includegraphics[width=0.81\linewidth]{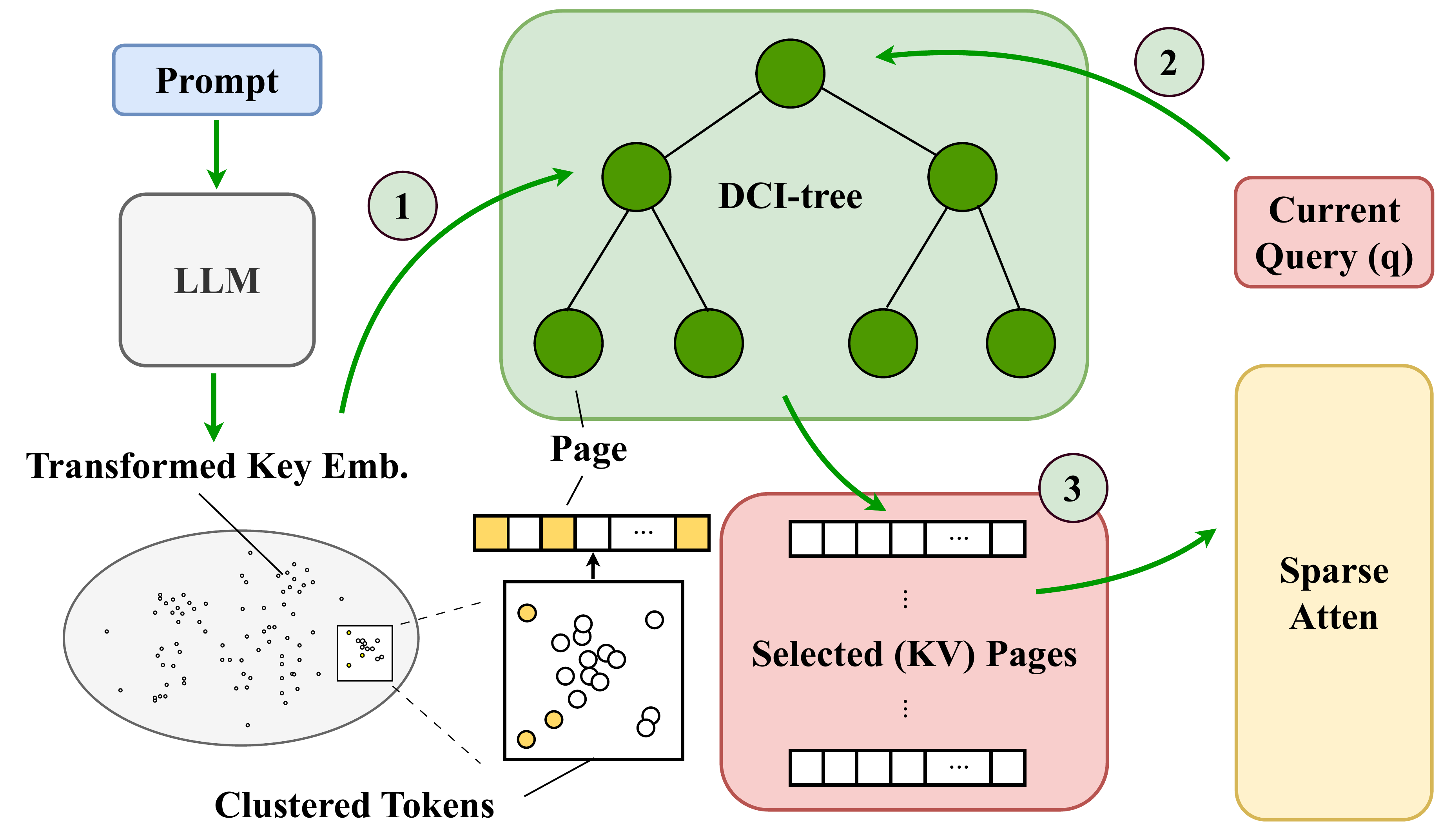}
\caption{\revise{Illustration of \toolName. (1) During the prefill stage, tokens are indexed into a hierarchical data structure (the DCI-tree) according to their semantic similarity in the transformed key-embedding space. Each leaf node of the DCI-tree corresponds to a physical memory page. (2) During the decoding stage, given a query $q$, \toolName performs a tree search to identify the top-$k$ tokens most relevant to $q$. The zoomed-in section at the bottom illustrates that these relevant tokens (highlighted in yellow) tend to be clustered within the same leaf nodes and are stored together in corresponding memory pages. (3) After the query-aware retrieval, the pages containing the relevant tokens are selected, and all tokens within these pages are used in the subsequent sparse attention with $q$.}}
\label{fig:icecache}
\end{figure}

To improve the identification of relevant KV-cache entries, we propose an innovative approach, \toolName, which integrates semantic token clustering with PagedAttention~\citep{kwon2023efficient}, a widely adopted memory management technique that stores KV-cache entries in non-contiguous pages. As illustrated in Figure~\ref{fig:icecache}, by grouping semantically related tokens into the same memory pages, our method increases the likelihood that relevant tokens are co-located, thereby improving page selection hit rates and enhancing memory bandwidth utilization during CPU–GPU transfers. Furthermore, \toolName employs a hierarchical data structure that can be efficiently updated during decoding, mitigating the performance degradation commonly observed in long-generation tasks. As a result, \toolName enables more effective and adaptive KV-cache management -- it achieves superior latency
and accuracy compared to state-of-the-art baselines under the same KV-cache budget, and still attains comparable accuracy while using substantially smaller budgets.
\vspace{-3mm}
In summary, \toolName makes the following contributions:
\vspace{-1mm}
\begin{itemize}
\item We propose a KV-cache offloading strategy that substantially improves retrieval hit rates by grouping semantically similar tokens into the same memory pages.
\item We develop a multi-level DCI algorithm that supports efficient indexing, fast retrieval, and dynamic updates, enabling effective maintenance of the KV-cache data structure.
\item We design a pipelining scheme that overlaps CPU and GPU computations, effectively hiding the latency associated with indexing and retrieval.
\end{itemize}

We evaluated \toolName under constrained GPU memory budgets on the Passkey Retrieval~\citep{mohtashami2023landmark}, LongBench~\citep{bai2023longbench}, and GSM8K Chain-of-Thought (CoT) reasoning~\citep{wei2022chain} using four popular open-source LLMs: Llama3.1-8B-Instruct, Mistral-7B-Instruct-v0.2, LongChat-7B-v1.5 and Qwen3-32B. Across diverse tasks, including open-domain QA, multi-hop reasoning, academic reading comprehension, long-context summarization and long-context generation, \toolName consistently outperformed six state-of-the-art KV-cache baselines. Notably, \toolName sustained near-oracle performance on most tasks using only a small fraction (as small as 64 tokens) of the original KV-cache size. Furthermore, by leveraging its hierarchical and dynamically updatable data structure, \toolName effectively mitigates performance degradation in long-generation tasks. As shown in Figure~\ref{fig:teaser}, the empirical results demonstrate that \toolName establishes a new state-of-the-art in accuracy–efficiency trade-offs, achieving superior accuracy while significantly reducing CUDA memory usage and decoding latency compared to all existing baselines.


\vspace{-3mm}
\section{Related work}
\label{sec:related_work}

\vspace{-1mm}
\subsection{KV-Cache Management via Eviction and Offloading}
\vspace{-2mm}
Existing approaches to KV-cache management generally fall into two categories: eviction-based methods and offloading-based methods. Eviction-based methods are typically faster, as they avoid data transfer between the GPU and the CPU. However, they often rely on static selection strategies and fail to adapt to changing contexts. In contrast, offloading-based methods aim to preserve important KV-cache entries more accurately by dynamically transferring data between GPU and CPU, but introduce additional latency. 

Eviction-based approaches include H2O~\citep{zhang2024h2o}, which retains only a subset of tokens selected based on attention scores; StreamingLLM~\citep{xiao2023efficient}, which preserves sink tokens and recent tokens to approximate attention; and SnapKV~\citep{li2024snapkv}, which selects key embeddings from the tail of the prompt to guide subsequent decoding. While efficient, these methods may struggle to maintain accuracy in long-generation scenarios due to their limited flexibility in adapting to evolving contexts. Offloading-based methods dynamically manage KV-cache storage across devices. MagicPiG~\citep{chen2024magicpig} uses sampling techniques based on Locality Sensitive Hashing (LSH) to approximate attention. OmniKV~\citep{hao2025omnikv} improves efficiency by reusing important tokens across consecutive layers. PQCache~\citep{zhang2024pqcache} applies product quantization to compress the KV-cache and approximate attention computation. Although these approaches better preserve critical information, they incur additional communication costs from GPU–CPU transfers.
\vspace{-1mm}
\subsection{Paged KV-Cache Management}
\vspace{-2mm}
PagedAttention~\citep{kwon2023efficient} is an innovative memory management technique designed to optimize the KV-cache of LLMs. It addresses the challenges by introducing a paging mechanism similar to virtual memory systems in operating systems. This approach divides the KV-cache into fixed-size pages, allowing for more efficient memory allocation and management. By doing so, PagedAttention improves GPU memory utilization by reducing fragmentation and enabling longer context windows without sacrificing performance. However, unlike the methods discussed in the previous subsection, PagedAttention does not address the fundamental issue that the KV-cache continuously grows during decoding.

Quest~\citep{tang2024quest} and ArkVale~\citep{chenarkvale} are two query-aware criticality estimation algorithms built on PagedAttention. They effectively identify relevant KV-cache tokens and perform self-attention selectively on the chosen tokens. For each page, Quest and ArkVale calculate an upper bound using the feature values of the Key vector for each page’s criticality estimation. Given all criticality scores of the pages, Top-$k$ pages are chosen to perform approximate self-attention, where $k$ is a preset constant (e.g. 128, 256). Additionally, ArkVale integrates the GPU-CPU offloading into the system to further save GPU memory. However, a key limitation of both Quest and ArkVale is that they construct memory pages according to the original token order. As a result, tokens that are semantically relevant to a given query may be scattered across multiple pages. In contrast, \toolName performs semantic clustering during indexing, grouping similar tokens into the same pages. This organization increases the likelihood that relevant tokens are co-located, improving retrieval hit rates and reducing unnecessary memory transfers.
\vspace{-3mm}
\section{Background}\label{sec:background}
\subsection{Attention Mechanism and Sparse Attention}
\vspace{-2mm}
Mathematically, the attention operation takes three matrices as input, $\mathbf{K} \in \mathbb{R}^{m \times d}, \mathbf{Q} \in \mathbb{R}^{n \times d}, \mathbf{V} \in \mathbb{R}^{m \times d'}$, which denote keys, queries and values, respectively. Optionally, it may also take in a mask as input, $\mathbf{S} \in \mathbb{R}^{n \times m}$, whose entries are either 0 or 1. The $i$th rows of $\mathbf{K}$, $\mathbf{Q}$ and $\mathbf{V}$, denoted as $\mathbf{k}_i$, $\mathbf{q}_i$ and $\mathbf{v}_i$, represent the $i$th key, query, and value, respectively. The entry of $\mathbf{S}$ in the $i$th row and $j$th column, denoted as $s_{i,j}$, represents whether the $i$th query is allowed to attend to the $j$th key --- if it is $1$, it would be allowed; if it is $0$, it would not be. A common masking scheme is the causal mask, where $s_{i,j}$ is $1$ if $i \geq j$ and $0$ otherwise. Keys and queries have the same dimension $d$, and each key is associated with a value, and so the number of keys and values is the same and denoted as $m$. The attention operation computes the attention weight matrix $\mathbf{A} \in \mathbb{R}^{n \times m}$. Its entry in the $i$th row and $j$th column, denoted as $a_{i,j}$, is computed with the following formula: 
\begin{equation}\label{approx_atten}
a_{i,j}=\frac{s_{i,j} \exp \left( \frac{\mathbf{q}_i^\top \mathbf{k}_{j}}{\sqrt{d}} \right)}{\sum_{j'=1}^{m} s_{i,j'} \exp \left( \frac{\mathbf{q}_i^\top \mathbf{k}_{j'}}{\sqrt{d}} \right)}
\end{equation}
The attention matrix $\mathbf{A}$ is typically sparse~\citep{nikita2020reformer, gupta2021memory}, i.e., in each row of $\mathbf{A}$, only a few attention weights have significant (large) values, while the majority of the remaining values are close to zero. If we can somehow identify the $k$ unmasked keys that receive the highest attention weights for each query $\mathbf{q}_i$ without computing the attention weights for all keys, the original attention matrix $\mathbf{A}$ can be approximated by only computing the inner product for the identified keys, which can save a significant amount of computational resources.
\vspace{-2mm}
\subsection{LLM Inference}
\vspace{-2mm}
The inference process of LLMs primarily comprises two key stages: the prefill (or prompt) stage and the decoding (or generation) stage. In the prefill stage, the model takes an input prompt sequence of length $s_{in}$ and processes it through all layers of the LLM. During this process, the keys and values for each token in the sequence are computed and stored as part of the KV-cache. The decoding stage begins once the prompt has been processed. Here, the model generates output tokens one step at a time, using and updating the KV-cache iteratively. For each decoding step, the current token's computation depends on the stored keys and values from previous tokens, allowing the model to maintain context over the sequence. The KV-cache thus plays a crucial role in enabling efficient autoregressive generation by reducing redundant computations and maintaining information about past tokens.
\vspace{-2mm}
\subsection{Multi-level DCI}\label{sec:dci}
\vspace{-2mm}
\noindent\textbf{Prioritized Dynamic Continuous Indexing (P-DCI).} \citet{li2016fast} proposed an exact, randomized algorithm designed to perform efficient $k$-nearest neighbour (k-NN) searches in high-dimensional spaces. Unlike traditional methods, DCI avoids space partitioning, which scales poorly to high dimensions. It does so by constructing multiple indices, each of which allows for points to be efficiently ranked by a lower bound on the distances between them and the query. The lower bound is formed by projecting the distances along a random vector associated with each index. P-DCI~\citep{li2017fast}  aggregates over the lower bounds given by each index to produce a tighter lower bound and uses a priority queue to order points so that points with smaller lower bounds are processed earlier than points with larger lower bounds. This approach significantly reduces the number of distance evaluations and memory usage compared to methods like Locality-Sensitive Hashing (LSH). \citet{mao2024iceformer} first introduced P-DCI into sparse attention mechanisms to accelerate LLM inference.

\textbf{Multi-level Dynamic Continuous Indexing (M-DCI).}  In this work, we extend P-DCI by introducing a hierarchical data structure to further enhance search efficiency. The index is organized into multiple levels, each containing a subset of data points. Points are randomly promoted to higher levels, forming a hierarchical data structure. Each point at a lower level is assigned a parent in the next higher level, typically the nearest neighbour among the promoted points. This creates "nodes" or clusters of points sharing the same parent. When querying, the algorithm starts at the top level, using P-DCI to find the $k$-closest points to the query. It then recursively searches within the nodes associated with these points at the next lower level, continuing this process down the hierarchy. This multi-level approach allows M-DCI to focus computational resources on the most promising regions of the index, effectively narrowing down the search space and improving query times, especially in datasets with high intrinsic dimensionality.

\section{\toolName}\label{sec:method}













\vspace{-3mm}
We propose an innovative approach, named \toolName, that integrates token clustering with KV-cache storage. Our method consists of three steps: (1) indexing; (2) page selection; and (3) bulk loading. The indexing step occurs either during the prompt processing phase—when \toolName constructs a hierarchical data structure, referred to as the DCI-tree, for the prompt key embeddings; or when new window pages are offloaded to the CPU. In this step, similar tokens are grouped into units called nodes, rather than being stored sequentially in virtual memory pages as in PagedAttention. Here, a node denotes a group of data points that share the same parent in the tree hierarchy. The next two steps take place during the token generation (decoding) phase. In the page selection step, \toolName employs a fast Approximate Nearest Neighbor (ANN) search algorithm, M-DCI, to independently select the top-$k$ most relevant pages for each attention head. Finally, in the bulk loading step, the selected pages are efficiently transferred from the CPU back to the GPU. IceCache also overlaps the indexing (a CPU-intensive operation) with ongoing GPU computations to further reduce the latency.

\begin{figure}[htb]
\centering
\includegraphics[width=0.96\linewidth]{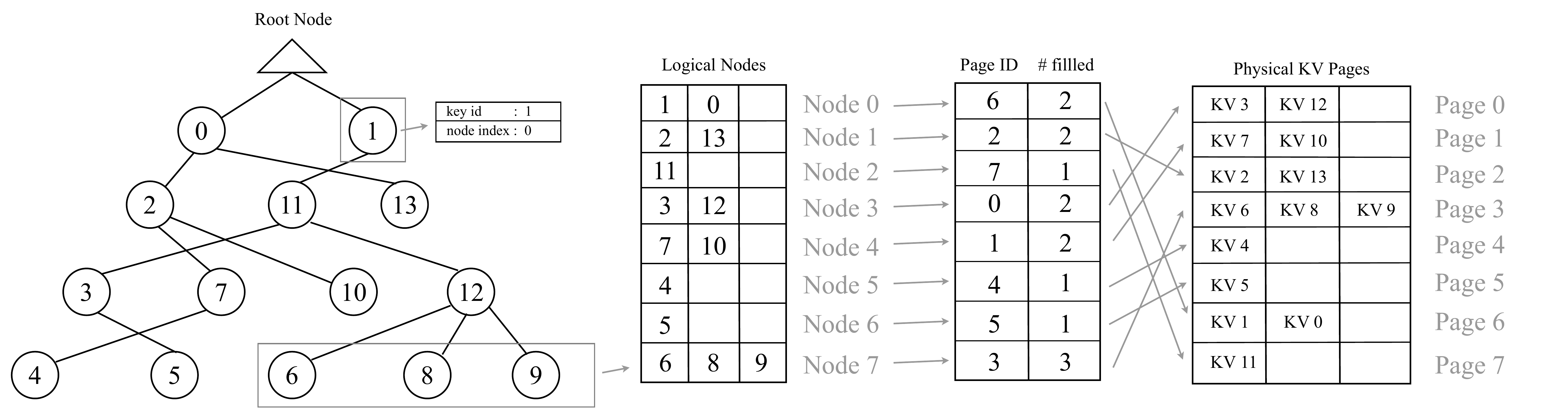}
  \caption{Illustration of DCI-tree and \toolName: The hierarchical data structure on the left visualizes the result of indexing key embeddings, DCI-tree, where each tree node stores metadata for the tokens, such as the key ID and node index. The tables on the right depict the mapping between nodes in the DCI-tree and the corresponding pages in physical memory. For each selected node, a mapping table is used to locate the memory region containing the associated key-value embeddings.}
  \label{fig:kvcache}
\end{figure}

We provide further details on each of these three steps in the following subsections and illustrate the method in Fig~\ref{fig:kvcache}.

\subsection{Indexing: Clustering Key Embeddings into a Hierarchical Tree}
During decoding, the dominant bottleneck is often not the computation but memory bandwidth. As the KV-cache grows with sequence length, each decoding step requires repeatedly loading large amounts of KV embeddings from GPU memory. This leads to memory-bandwidth saturation, significantly degrading latency scalability. PagedAttention~\citep{kwon2023efficient} mitigates this issue by introducing a virtual memory abstraction for the KV-cache. Instead of storing the KV-cache as a single contiguous tensor, it organizes KVs into fixed-size memory pages and maintains a page table to map logical token positions to physical memory locations. 
Within each page, consecutive tokens are stored contiguously, reducing fragmentation. 
This paging mechanism enables more flexible memory allocation and more efficient KV-cache management during decoding.

Building on this design, subsequent KV-cache optimization methods such as Quest~\citep{tang2024quest} and ArkVale~\citep{chenarkvale} leverage PagedAttention’s paging abstraction while incorporating query-aware page selection. These approaches estimate the importance of each page during decoding and utilize only the top-$k$ pages to approximate attention more efficiently.

However, since pages are constructed based on the original token order, tokens relevant to a given query are often distributed across multiple pages. Therefore, retrieving them requires loading entire pages that may contain many irrelevant tokens, leading to unnecessary memory overhead and reduced selection precision. In contrast, \toolName organizes KV-cache entries based on semantic similarity, concentrating relevant tokens within fewer pages. This design improves retrieval efficiency and achieves superior performance while reducing the number of pages accessed.

Specifically, \toolName organizes key-value embeddings into memory pages through a fundamentally different indexing strategy: instead of relying on the token's original order, \toolName constructs a separate hierarchical tree data structure for each attention head, called a DCI-tree, which clusters tokens based on the semantic similarity of their key embeddings. Each node in the DCI-tree represents a small group of semantically related tokens that share a common parent, effectively forming a localized cluster. From a memory system perspective, \toolName maps each node directly to a memory page, thereby preserving semantic locality in storage and enabling efficient access during decoding. By clustering semantically similar tokens into the same nodes/pages, \toolName enables more targeted and efficient retrieval during decoding. 

Moreover, the DCI-tree data structure used by \toolName is designed for efficient incremental updates. As new windows of tokens (e.g., from a sliding window in long-context scenarios) are offloaded to the CPU, each token is inserted into the appropriate node in the DCI-tree based on its key embedding. When a node exceeds the maximum page size, new pages are dynamically allocated to maintain balance. This adaptive tree maintenance ensures that the index remains both semantically meaningful and efficient, making \toolName particularly effective for long-sequence generation.

In summary, while prior methods treat KV-cache page layout as a static memory allocation problem, \toolName introduces a dynamic, semantically-aware data structure that preserves key similarity across time. This enables more focused page retrieval, reduces memory fragmentation, and supports more efficient decoding. Furthermore, by performing indexing during the prompt or CPU offloading phase, \toolName amortizes the tree construction cost and avoids incurring additional latency during inference.

\subsection{Indexing Details}\label{sec:indexing}
For each attention head, given a set of pre-computed key embeddings, \toolName first indexes them using a hierarchical tree structure which is obtained by a novel approach called Multi-level DCI (M-DCI). It works by constructing a dynamic index called DCI-tree and applies Prioritized DCI (P-DCI)~\citep{li2017fast} to each level of the tree recursively (more details are in Section~\ref{sec:dci}). The data points processed in M-DCI are transformed key embeddings and query embeddings using the following transformation formulas, which we denote as $T_K: \mathbb{R}^d \rightarrow \mathbb{R}^{d+1}$ and $T_Q: \mathbb{R}^d \rightarrow \mathbb{R}^{d+1}$:
\begin{align}
T_K(\mathbf{k}_j) & =\begin{bmatrix}\mathbf{k}_j / c & 
\sqrt{1-\|\mathbf{k}_j\|_2^2 / c^2}\end{bmatrix}^\top \\
T_Q(\mathbf{q}_i) & =\begin{bmatrix}
\mathbf{q}_i / \|\mathbf{q}_i\|_2 &
0
\end{bmatrix}^\top
\end{align}
where $c \geq \max _{j'}\|\mathbf{k}_{j'}\|_2$ is at least the maximum norm across all keys. We use the Euclidean distance as the distance function. 

At the very beginning of the indexing, all data points are initially placed at the bottom level of the DCI-tree. Subsequently, some points are randomly selected to be promoted to the next higher level based on a promotion ratio $r < 1$. The ratio $r$ is predefined during DCI-tree initialization and remains fixed throughout the process. After the indexing, we can get the total number of levels in the DCI-tree, denoted as $L$. The details are presented in Algorithm~\ref{alg_prompt}.

Specifically, let $n_\ell$ denote the number of data points at level $\ell$, with level indices starting from the bottom (i.e., the lowest level is $\ell = 1$). Ideally, the number of points satisfies the recurrence relation $n_\ell = r \cdot n_{\ell-1}$. In other words, the distribution over level indices follows a geometric distribution. The probability that a point is assigned to the highest level ($\ell = L$) is $r^{L-1}$, while the probability of being assigned to level $\ell$ (for $1 \leq \ell \leq L-1$) is $r^{\ell-1} - r^{\ell}$.

After level assignment, each data point at level $\ell$ is linked to a parent at level $\ell + 1$, defined as the closest point in terms of key embedding distance. This parent assignment is formulated as a 1-nearest neighbor search and is efficiently solved using M-DCI query.

In the decoding stage, when a new token is generated, its key embedding is inserted into the appropriate position in the DCI-tree. A level $\ell$ is first assigned to the new key according to the same random promotion process. Then, its parent at level $\ell + 1$ is determined, and the key is added to the physical memory page corresponding to the node into which it is inserted.



\subsection{Page Selection: Head-specific ANN Search with Fine-grained Retrieval}
\label{sec:page_selection}
\vspace{-2mm}
During the decoding phase, given a query embedding, \toolName performs head-specific page selection to identify the most relevant pages for each attention head. Leveraging the hierarchical DCI-tree constructed during indexing, we apply the fast approximate nearest neighbor (ANN) search algorithm described in Section~\ref{sec:dci}, M-DCI, to retrieve the top-$k$ pages that are important to the current query for each head independently. For group-query attention (GQA), where multiple query heads share the same set of key embeddings, we compute the union of the selected pages across queries within the same group and share the resulting pages among them, following~\citet{yuan2025native}.

\subsection{Page Selection Details}\label{sec:select}
As aforementioned, \toolName aims to accelerate self-attention by loading only a limited number of pages into GPU memory for computation. Therefore, the objective of page selection is to maximize the \emph{recall} (or hit rate) of significant keys for a given query. By clustering semantically similar tokens into the same nodes/pages, \toolName enables more targeted and efficient retrieval during decoding. In contrast, methods like Quest~\citep{tang2024quest}, Arkvale~\citep{chenarkvale}, or PQCache~\citep{zhang2024pqcache} construct pages based on the original token order, which often causes tokens relevant to a given query to be scattered across multiple pages. Retrieving them requires loading entire pages filled with many irrelevant tokens, resulting in unnecessary memory overhead. \toolName mitigates this inefficiency by grouping similar tokens, so relevant tokens tend to be concentrated within fewer pages. As a result, the hit rate of significant keys during decoding increases. The detailed procedure is shown in Algorithms~\ref{alg_page_select} and~\ref{alg_query}.

Specifically, when computing the attention matrix, given a query vector $q_i$, we follow the query process described in Section~\ref{sec:dci} to identify the top-$k$ keys that are most likely to yield the highest dot-product values with $q_i$. Once these top-$k$ keys are identified, we load only the pages that contain them. Suppose $p$ pages are loaded, and each page contains $d$ entries, since not all the pages are fully filled, the number of loaded keys $N$ is bounded as: $N \leq pd$.

The approximate attention scores between the query $q$ and these $N$ selected keys are then computed using Equation~\ref{approx_atten}. The masks \( s_{i,j} \) are set to 1 for the selected keys and 0 for all others. Note that, \toolName constructs a separate DCI-tree for each attention head, allowing it to retrieve different sets of significant pages per head. This head-specific, fine-grained selection mechanism distinguishes \toolName from baselines such as Quest and ArkVale, which retrieve the same set of pages for all heads, potentially limiting their retrieval accuracy.

\vspace{-2mm}
\subsection{Bulk Loading and Pipelining}
\vspace{-2mm}
In this section, we present how we optimize the efficiency of the \toolName workflow, using bulk loading and pipelining. 
Our key observation is that the selected pages are not contiguous in either main memory or GPU memory.
As a result, individual transfer of these pages between main memory and GPU memory is highly inefficient.
To overcome this issue, we designed bulk loading algorithms to efficiently offload and backload the selected pages, using two CPU and GPU backload buffers.
In addition, we carefully designed an efficient pipeline of prefilling calculations, KV-cache offloading, and DCI indexing, for efficient prefilling.
\vspace{-12mm}
\begin{figure}[!htb]
  \centering
  \begin{minipage}[c]{0.48\linewidth}
    \centering
    \vspace{0.7\baselineskip}
    \includegraphics[width=\linewidth]{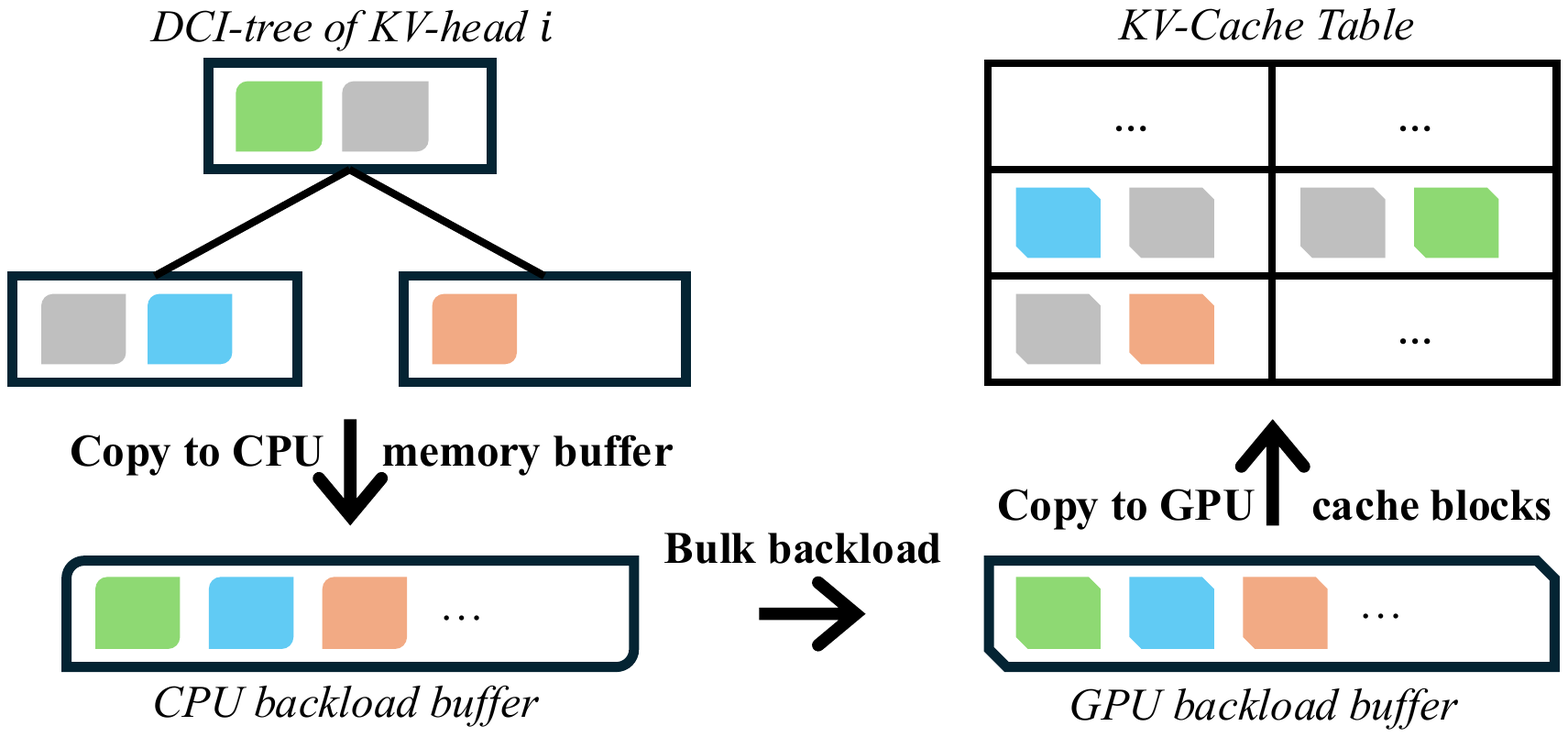}

    \vspace{-6mm}
    \caption{\revise{After \toolName selects important KV-pages, it aggregates all selected pages into a contiguous CPU preloading buffer. This buffer is then transferred via high-throughput PCIe transaction to a pre-allocated GPU buffer. Finally, the transferred blocks are scattered into their exact locations in the KV-Cache table. This bulk transfer avoids many small copying operations over PCIe and significantly improves utilization.}
    }
    \label{fig:bulk-backload}
  \end{minipage}
  \hspace{0.015\linewidth}
  \begin{minipage}[c]{0.48\linewidth}
  \vspace{7mm}
    \centering
    \includegraphics[width=\linewidth]{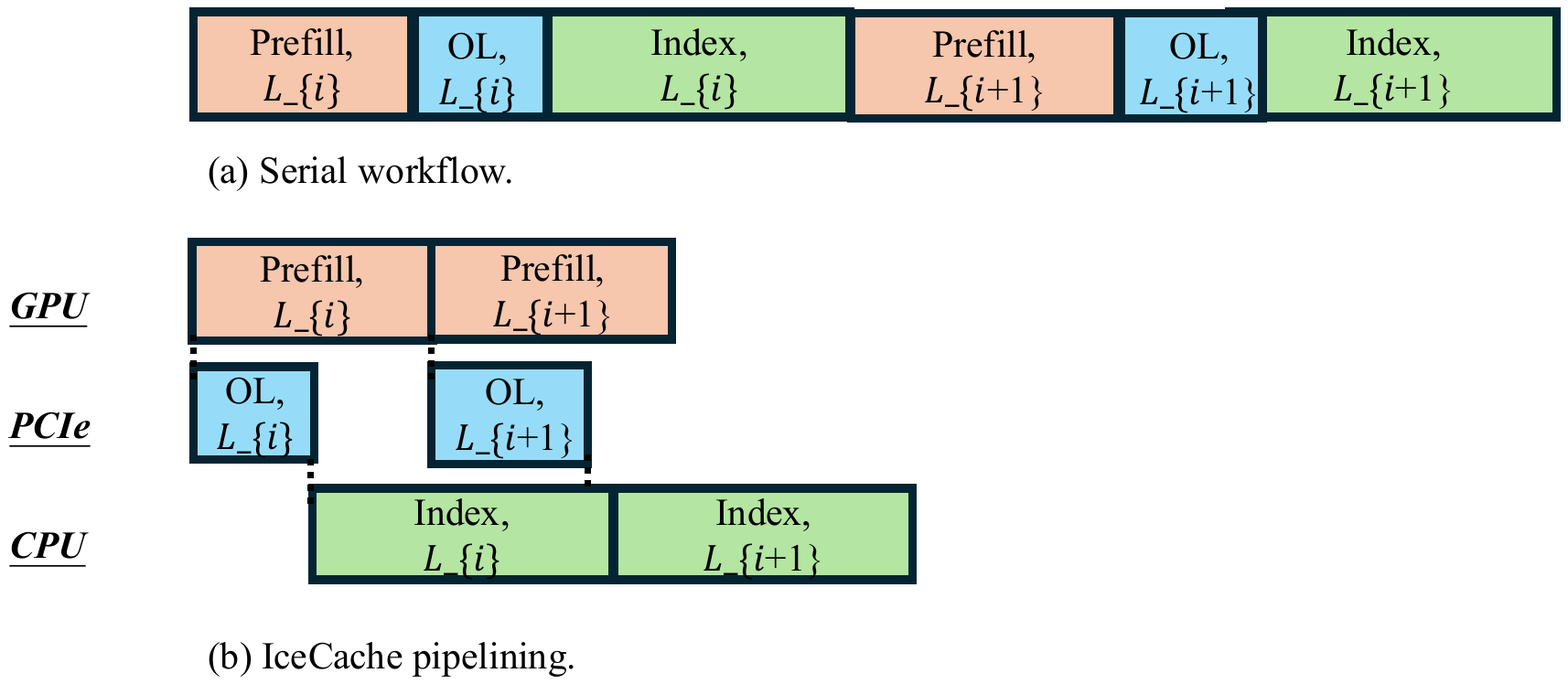} 
    \vspace{-11mm}
    \caption{
    \revise{(a) Baseline serial workflow, where prefilling, offloading (OL), and indexing are executed strictly in sequence. (b) IceCache pipelining, where GPU prefilling overlaps with KV-offloading via PCIe and CPU-side DCI indexing. Once KVs of layer $i$ ($L_{i}$) arrive in CPU memory, Li-DCI-tree indexing progresses in parallel with GPU prefilling and offloading of the subsequent layer ($L_{i+1}$). This results in significantly reduced end-to-end prefilling latency.}
    }
    \label{fig:backload-pipeline}
  \end{minipage}
\end{figure}
\vspace{-2mm}
We illustrate our bulk back-loading workflow in Figure \ref{fig:bulk-backload}.
Offloading follows a similar procedure, except that the steps are executed in reverse.
After identifying the most relevant pages (indicated by color), we filter out those already resident in GPU memory from previous token generations. 
The remaining pages are then aggregated into a pre-allocated CPU‐memory buffer, enabling a single high-throughput PCIe transfer into a pre-allocated GPU‐memory buffer. 
Once the pages arrive in the GPU, we perform a scatter operation to load them into the appropriate slots of the KV-Cache table.

Figure~\ref{fig:backload-pipeline}(b) illustrates the pipelining of \toolName prefilling (other minor stages are omitted for clarity).
After the KV embeddings are generated on the GPU, we simultaneously launch the prefilling calculation and the offloading transfer. 
The DCI index construction starts building the index along with the prefilling calculation, right after the KV states fully arrive in the main memory. 
This approach allows the offloading and indexing latencies to be largely hidden by the main prefilling computation. 
Furthermore, \toolName can be easily extended with page reuse techniques~\citep{liu2023deja,hao2025omnikv} to further accelerate prefilling.

\vspace{-2mm}
\section{Experiments} \label{sec:exp}
  \vspace{-2mm}
\subsection{Settings}
\vspace{-2mm}
To comprehensively evaluate \toolName, we conduct experiments across models of varying scales and attention architectures. We first apply \toolName to Llama-3.1-8B-Instruct and Mistral-7B-Instruct-v0.2, two widely adopted open-source LLMs employing group-query attention (GQA)~\citep{ainslie2023gqa}. To further assess scalability and architectural generality, we extend our evaluation to Qwen3-32B, a larger-scale model, and LongChat-7B-v1.5, which employs standard multi-head attention~\citep{vaswani2017attention} rather than group-query attention. Our evaluation proceeds in four stages. We first measure recall in retrieving important tokens~\citep{mohtashami2023landmark}. To evaluate the performance of \toolName on long generation tasks, we further conduct experiments on LongBench~\citep{bai2023longbench}, which contains long generation tasks such as summarization and code generation, and GSM8K~\citep{cobbe2021training}, which requires chain-of-thought reasoning. Finally, we examine its scalability on RULER~\citep{hsieh2024ruler} under extremely long-context settings.


Our experimental platform comprises an NVIDIA A100 40GB PCIe GPU (for small models) or an NVIDIA H100 80GB PCIe GPU (for large models). For all experiments, we fix the number of CPU threads to 64. The software stack includes CUDA version 12.2, PyTorch version 2.4.1, and HuggingFace Transformers version 4.57.1. We implement \toolName on top of HuggingFace Transformers, utilizing FlashInfer~\citep{ye2025flashinfer} for the attention kernel operation. As prior research indicates~\citep{tang2024quest}, the initial layers of the model exhibit relatively low sparsity. Therefore, neither \toolName nor baseline methods are applied to the first two layers of the models.

\vspace{-2mm}
\subsection{Passkey Retrieval Accuracy}
\begin{figure}[htb]
\centering
\begin{subfigure}[t]{0.32\linewidth}
    \centering
    \includegraphics[width=\linewidth]{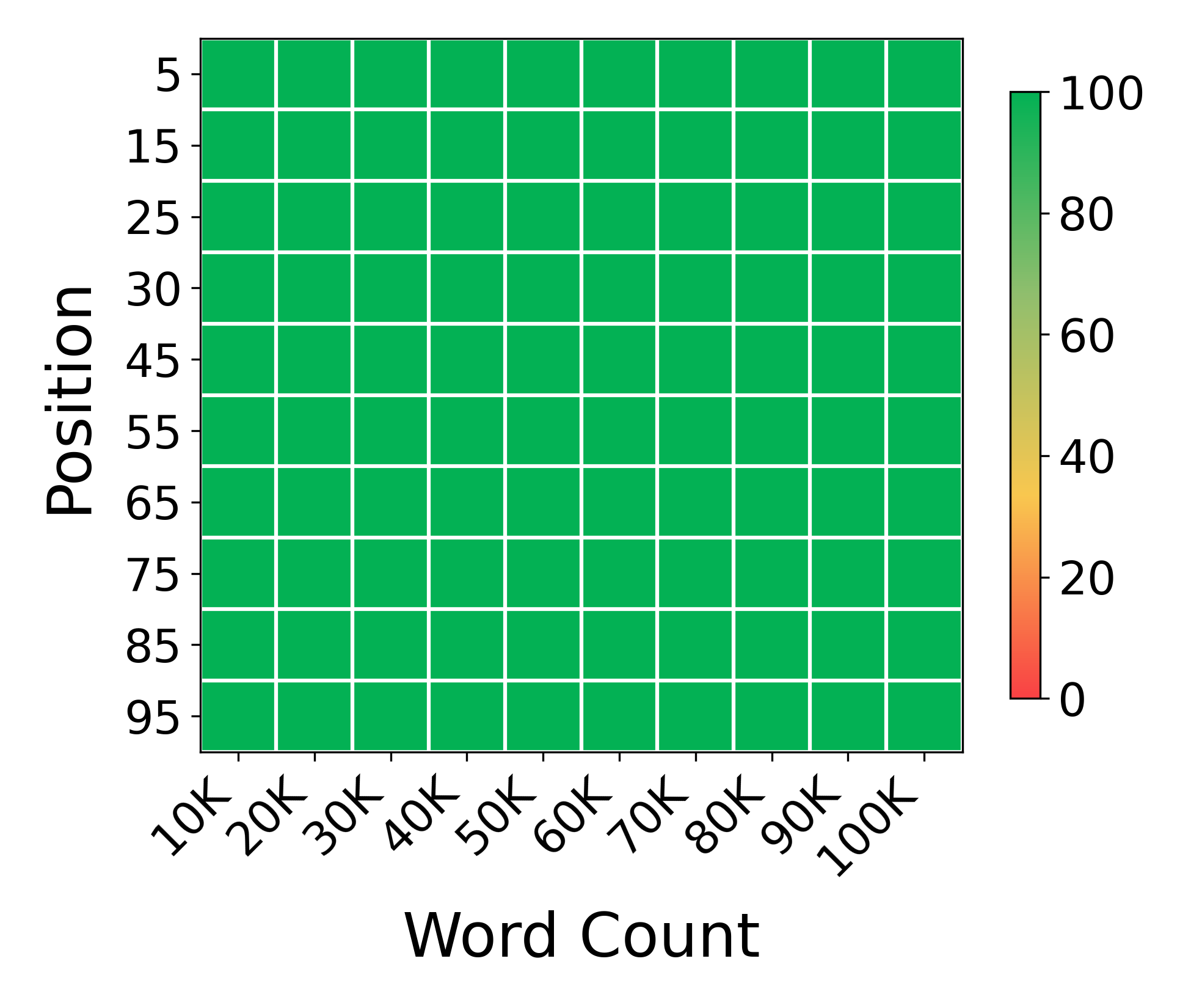}
    \caption{Cache Budget = 256}
\end{subfigure}
\begin{subfigure}[t]{0.32\linewidth}
    \centering
    \includegraphics[width=\linewidth]{pass_heatmap.png}
    \caption{Cache Budget = 128}
\end{subfigure}
\begin{subfigure}[t]{0.32\linewidth}
    \centering
    \includegraphics[width=\linewidth]{pass_heatmap.png}
    \caption{Cache Budget = 64}
\end{subfigure}
\caption{
Passkey retrieval accuracy of \toolName on Llama3.1-8B-Instruct. The horizontal axis indicates the relative insertion position (\%) of the passkey, while the vertical axis represents the context length in words. Results are presented for cache budgets of 256, 128, and 64. Notably, \toolName achieves 100\% retrieval accuracy across all tested budget sizes.
}
\label{fig:passkey}
\end{figure}

\vspace{-2mm}
We first evaluate \toolName's effectiveness in handling long-range dependencies using the passkey retrieval task. We consider context lengths from 10k words to 100k words, and test with the size of cache budget $=\{256, 128, 64\}$. For each length, 100 test cases are generated with passkeys inserted at various positions from 0\% to 95\% of the total context length in increments of 5\%. The results are illustrated in Fig.~\ref{fig:passkey}. As shown in the figure, \toolName dynamically assesses the importance of evicted pages and recall crucial ones on demand, consistently maintaining 100\% retrieval accuracy across all tested budget sizes.

\vspace{-2mm}
\subsection{LongBench Evaluation
}
\vspace{-2mm}
To assess the performance of our method in long-context scenarios, we conduct a comprehensive evaluation on the LongBench benchmark~\citep{bai2023longbench}. LongBench is designed to evaluate how well LLMs understand and reason over long-context inputs across diverse real-world tasks. We compare \toolName (ICE) against six state-of-the-art KV cache optimization methods, including MagicPig (MPG)~\citep{chen2024magicpig}, ArkVale (AKV)~\cite{chenarkvale}, SnapKV (SKV)~\cite{li2024snapkv}, StreamingLLM (SLM)~\cite{tang2024quest}, OmniKV (OKV)~\cite{hao2025omnikv} and PQCache (PQC)~\cite{zhang2024pqcache} as baselines. We also include the results of full KV-cache (FULL) and ground-truth top-$k$ KV-cache (TOP-$k$) as baselines. Notably, OmniKV offloads only a subset of layers while keeping all remaining layers fully resident on the GPU, resulting in substantially higher memory usage than other methods. The detailed results are presented in Table \ref{table:longbench}.  

\vspace{-2mm}
\subsubsection{Accuracy Analysis}
\vspace{-2mm}
\paragraph{Accuracy on Llama-3.1-8B-Instruct.}
On Llama-3.1-8B, the effectiveness of \toolName is particularly pronounced. Most impressively, with a highly constrained KV-Cache budget of just 64, our method achieves an average accuracy of 47.8. This result alone surpasses the strongest baseline, PQCache, which scores 47.3 while operating with a 4$\times$ larger budget of 256. This highlights the exceptional efficiency of our approach in low-resource environments. As we increase the budget for \toolName to 256, its performance climbs to an average score of 49.0. This not only represents a substantial 1.7 point improvement over PQCache but also closes the performance gap to the unconstrained Full KV-Cache (49.5) to a mere 0.5 points. Notably, our performance is remarkably close to the ground-truth Top-$k$, demonstrating a near-optimal cache management strategy across a diverse set of tasks.
\vspace{-3mm}
\paragraph{Accuracy on Mistral-7B-Instruct.}
This strong performance trend is consistent on the Mistral-7B model, confirming the robustness of our method. With a budget of 256, \toolName achieves an average accuracy of 41.7, establishing a significant 2.6 point lead over the best-performing baseline, MagicPig (39.1). Again, the low-budget capability of \toolName stands out; with a budget of 64, \toolName scores 39.0, remaining highly competitive with the top baseline (MagicPig, scores 39.1) that uses four times the cache size (256).
\begin{table*}[t]
\centering
\caption{Accuracy comparison of our method (ICE) with SnapKV (SKV), SteamingLLM (SLM), OmniKV (OKV), MagicPig (MPG), PQCache (PQC), ArkVale (AKV), Full KV (FULL) and ground-truth top-$k$ (TOP-$k$) on LongBench for Llama-3.1-8B-Instruct and Mistral-7B-Instruct. \toolName generally outperforms other methods across various KV-cache budgets and LLMs.}
\footnotesize  
\setlength{\tabcolsep}{2pt}  
\scalebox{0.612}{
\begin{tabular}{c|c|c|c|c|c|c|c|c|c|c|c|c|c|c|c|c|c|c}
\midrule
\multirow{3}{*}{\textbf{Budget}} & \multirow{3}{*}{\textbf{Method}} & \multicolumn{3}{c|}{\textbf{Single-Document QA}} & \multicolumn{3}{c|}{\textbf{Multi-Document QA}} & \multicolumn{3}{c|}{\textbf{Summarization}} & \multicolumn{3}{c|}{\textbf{Few-shot Learning}} & \multicolumn{2}{c|}{\textbf{Synthetic}} & \multicolumn{2}{c|}{\textbf{Code}} & \multirow{3}{*}{\textbf{Avg.}} \\
\cmidrule{3-18}
& & NrtvQA & Qasper & MF-en & HotpotQA & 2WikiMQA & Musique & GovReport & QMSum & MultiNews & TREC & TriviaQA & SAMSum & PCount & PRe & Lcc & RB-P & \\
& & 18409 & 3619 & 4559 & 9151 & 4887 & 11214 & 8734 & 10614 & 2113 & 5177 & 8209 & 6258 & 11141 & 9289 & 1235 & 4206 & \\
\midrule
\midrule
\multicolumn{19}{c}{\textbf{Llama-3.1-8B-Instruct}} \\
\midrule
N/A & FULL & 30.2 & 45.5 & 54.9 & 55.5 & 46.7 & 31.3 & 35.2 & 25.2 & 27.2 & 72.5 & 91.7 & 43.8 & 8.4 & 99.5 & 65.1 & 58.8 & 49.5 \\
256 & TOP-{$k$} & 30.7 & 44.7 & 55.4 & 55.0 & 46.5 & 31.7 & 34.8 & 25.1 & 26.8 & 71.5 & 92.2 & 44.8 & 7.8 & 100.0 & 67.1 & 63.6 & 49.8 \\
\midrule
\multirow{6}{*}{256} & SKV & 23.7 & 27.3 & 46.3 & 52.3 & 40.8 & 24.2 & 19.2 & 22.6 & 19.2 & 29.0 & 84.1 & 39.0 & \textbf{8.6} & 97.5 & 57.3 & 56.2 & 40.8 \\
& SLM & 17.0 & 23.2 & 25.7 & 21.0 & 29.3 & 6.8 & 20.8 & 17.5 & 20.7 & 45.5 & 84.3 & 41.2 & 5.0 & 71.5 & 59.5 & 47.5 & 33.5 \\
& OKV & 27.2 & 40.7 & 52.8 & 55.1 & 45.6 & 29.7 & 27.6 & 23.0 & 25.4 & 72.5 & 88.9 & 40.4 & 5.6 & 94.5 & 60.8 & 51.5 & 46.3 \\
& MPG & 25.5 & 39.9 & 51.8 & 51.4 & 39.5 & 25.7 & 33.9 & 23.5 & 25.9 & 65.5 & 84.0 & 37.0 & 7.8 & 99.5 & 47.4 & 44.8 & 44.6 \\
& PQC & 28.7 & 43.3 & 52.2 & 55.2 & 45.1 & 28.4 & 27.2 & 24.0 & 22.8 & 69.5 & 91.1 & 41.2 & 6.2 & 99.0 & 59.0 & 54.4 & 47.3 \\
& AKV & 26.1 & 35.2 & 47.5 & 51.6 & 45.6 & 28.1 & 22.9 & 22.5 & 22.8 & 53.5 & 90.1 & 40.0 & 6.7 & 85.0 & 57.7 & 48.9 & 42.8 \\
\midrule
\rowcolor{green!10}
64 & ICE & 27.4 & 43.2 & 55.7 & \textbf{55.3} & 44.4 & \textbf{31.2} & 33.4 & 23.7 & 26.2 & 72.5 & 90.3 & 41.9 & 6.6 & 99.5 & 61.7 & 51.6 & 47.8 \\
\rowcolor{green!10}
128 & ICE & 30.0 & \textbf{44.7} & \textbf{56.5} & 55.0 & 45.4 & 30.0 & 33.5 & 24.3 & 26.5 & \textbf{73.0} & 91.3 & 42.4 & 6.5 & \textbf{100.0} & 61.5 & 56.7 & 48.6 \\
\rowcolor{green!10}
256 & ICE & \textbf{30.6} & \textbf{44.7} & 56.3 & 55.2 & \textbf{45.9} & 30.6 & \textbf{34.6} & \textbf{24.4} & \textbf{26.7} & \textbf{73.0} & \textbf{92.0} & \textbf{43.5} & 6.7 & \textbf{100.0} & \textbf{62.5} & \textbf{56.4} &\textbf{49.0} \\
\midrule
\midrule
\multicolumn{19}{c}{\textbf{Mistral-7B-Instruct-v0.2}} \\
\midrule
N/A & FULL & 26.8 & 33.1 & 49.3 & 42.8 & 27.3 & 18.8 & 33.0 & 24.2 & 27.1 & 71.0 & 86.2 & 42.8 & 2.9 & 87.0 & 56.9 & 54.3 & 42.2 \\
256 & TOP-{$k$} & 26.2 & 31.3 & 48.9 & 40.0 & 26.2 & 19.6 & 33.3 & 24.2 & 27.1 & 73.0 & 86.6 & 43.3 & 2.3 & 82.9 & 59.0 & 57.3 & 42.6 \\
\midrule
\multirow{6}{*}{256} & SKV & 18.3 & 15.1 & 38.3 & 30.4 & 19.2 & 13.8 & 16.5 & 21.4 & 19.9 & 32.0 & 81.7 & 38.1 & \textbf{4.0} & 59.1 & 49.0 & 51.8 & 31.8 \\
& SLM & 14.2 & 12.3 & 26.4 & 23.2 & 14.8 & 10.1 & 17.5 & 19.8 & 19.8 & 51.0 & 80.5 & 39.9 & \textbf{4.0}& 15.6 & 52.0 & 45.4 & 27.8 \\
& OKV & 16.5 & 22.8 & 43.8 & 34.7 & 19.2 & 16.6 & 24.4 & 22.0 & 24.8 & 70.5 & 81.7 & 38.6 & 3.1 & 35.2 & 36.5 & 38.4 & 33.0 \\
& MPG & 21.0 & {28.0} & 46.5 & 38.4 & 19.5 & 17.5 & 30.8 & \textbf{23.3} & 25.8 & 70.0 & 83.7 & 39.5 & 2.5 & 85.0 & 49.4 & 48.3 & 39.1 \\
& PQC & 21.0 & 25.8 & 42.5 & 18.1 & 20.3 & 16.0 & 29.5 & 21.7 & \textbf{27.1} & 70.0 & 85.8 & 39.7 & 2.5 & 75.5 & 54.2 & 49.2 & 37.4 \\
& AKV & 18.0 & 15.2 & 41.5 & 30.7 & 17.0 & 13.2 & 21.6 & 21.6 & 23.1 & 58.0 & 85.8 & 40.6 & 2.1 & 41.5 & 51.4 & 47.2 & 33.0 \\
\midrule
\rowcolor{green!10}
64 & ICE & 23.9 & 29.0 & 47.4 & \textbf{40.5} & 23.5 & 18.3 & 30.6 & 21.8 & 26.0 & 70.5 & 85.9 & 41.5 & 3.5 & 56.5 & 53.2 & 50.4 & 39.0 \\
\rowcolor{green!10}
{128} & ICE & 25.1 & 30.5 & \textbf{48.7} & 40.4 & \textbf{26.7} & \textbf{18.7} & 30.3 & 22.0 & 26.1 & 70.5 & 85.7 & 42.4 & 3.4 & 70.2 & 53.6 & 51.5 & 40.4 \\
\rowcolor{green!10}
256 & ICE & \textbf{25.2} & \textbf{31.0} & {48.4} &{40.4} & 26.0 & 18.3 & \textbf{31.5} & 23.1 & 26.9 & \textbf{71.0} & \textbf{86.3} & \textbf{42.8} & 3.9 & \textbf{85.1} & \textbf{54.7} & \textbf{53.2} & \textbf{41.7}\\
\midrule
\end{tabular}
}
\label{table:longbench}
\end{table*}



\begin{table*}[t]
\centering
\caption{Accuracy comparison of our method (ICE) with Full KV (FULL) on LongBench for Qwen3-32B and LongChat-7B-v1.5.}
\footnotesize  
\setlength{\tabcolsep}{2pt}  
\scalebox{0.612}{
\begin{tabular}{c|c|c|c|c|c|c|c|c|c|c|c|c|c|c|c|c|c|c}
\midrule
\multirow{3}{*}{\textbf{Budget}} & \multirow{3}{*}{\textbf{Method}} & \multicolumn{3}{c|}{\textbf{Single-Document QA}} & \multicolumn{3}{c|}{\textbf{Multi-Document QA}} & \multicolumn{3}{c|}{\textbf{Summarization}} & \multicolumn{3}{c|}{\textbf{Few-shot Learning}} & \multicolumn{2}{c|}{\textbf{Synthetic}} & \multicolumn{2}{c|}{\textbf{Code}} & \multirow{3}{*}{\textbf{Avg.}} \\
\cmidrule{3-18}
& & NrtvQA & Qasper & MF-en & HotpotQA & 2WikiMQA & Musique & GovReport & QMSum & MultiNews & TREC & TriviaQA & SAMSum & PCount & PRe & Lcc & RB-P & \\
& & 18409 & 3619 & 4559 & 9151 & 4887 & 11214 & 8734 & 10614 & 2113 & 5177 & 8209 & 6258 & 11141 & 9289 & 1235 & 4206 & \\
\midrule
\midrule
\multicolumn{19}{c}{\textbf{Qwen3-32B}} \\
\midrule
N/A & FULL & 32.6 & 45.5 & 50.4 & 59.6 & 56.0 & 40.1 & 33.2 & 23.9 & 25.2 & 72.0 & 70.8 & 37.1 & 18.0 & 100.0 & 12.4 & 18.4 & 43.4 \\
\midrule
64 & ICE & 29.5 & 43.9 & 50.9 & 61.4 & 55.8 & 38.8 & 30.8 & 23.8 & 24.3 & 71.0 & 70.0 & 35.7 & 15.0 & 97.0 & 10.5 & 17.3 & 42.2 \\
128 & ICE & 32.1 & 44.8 & 53.3 & 61.6 & 54.6 & 38.5 & 31.0 & 23.6 & 24.8 & 72.0 & 70.3 & 36.2 & 15.5 & 100.0 & 10.7 & 17.3 & 42.6  \\
256 & ICE & 32.2 & 44.1 & 51.9 & 60.2 & 55.1 & 38.9 & 32.4 & 24.3 & 24.7 & 71.5 & 70.7 & 37.4 & 16.5 & 100.0 & 11.8 & 18.0 & 43.1 \\
\midrule
\midrule
\multicolumn{19}{c}{\textbf{LongChat-7B-v1.5}} \\
\midrule
N/A & FULL & 20.8 & 29.4 & 43.1 & 33.0 & 24.4 & 14.7 & 30.8 & 22.8 & 26.7 & 66.5 & 84.0 & 22.5 & 0.0 & 30.5 & 54.7 & 59.2 & 35.2 \\
\midrule
64 & ICE & {18.5} & {26.9} & {40.6} & {34.2} & {22.7} & {14.0} & {29.3} & {20.9} & {25.6} & {66.5} & {84.3} & {21.9} & {1.5} & {26.1} & {52.5} & {56.8} & 33.9\\
{128} & ICE & {19.9} & {27.7} & {40.9} & {33.9} & {24.2} & {14.4} & {28.6} & {21.8} & {25.9} & {66.5} & {84.2} & {22.3} & {1.5} & {27.3} & {52.6} & {57.7} & 34.3 \\
256 & ICE & {20.4} & {29.5} & {43.0} & {34.6} & {23.7} & {14.2} & {29.8} & {22.7} & {26.1} & {66.5} & {84.7} & {22.9} & {0.0} & {28.5} & {53.6} & {59.0} & 35.0 \\
\midrule
\end{tabular}
}
\label{table:two_models}
\end{table*}
\vspace{-3mm}
\paragraph{Accuracy on two additional LLMs.} To further validate \toolName on larger-scale models and standard multi-head attention architectures, we evaluate it on LongBench using two additional models: Qwen3-32B and LongChat-7B-v1.5. As shown in Tables~\ref{table:two_models}, for Qwen3-32B, IceCache with a small budget of 64 achieves an average accuracy of 42.2 on LongBench, retaining 97.2\% of the full KV-cache performance (43.4). This score rises to 99.3\% with a budget of 256, nearly matching the vanilla model. Similarly, on LongChat-7B-v1.5, our method preserves 96.3\% of the full KV-cache performance with a budget of 64, and achieves up to 99.4\% at a budget of 256. These results provide strong evidence that \toolName is effective across different model scales and attention mechanisms.
\vspace{-2mm}
\subsubsection{Latency Analysis}
\begin{figure}[!htb]
\centering
\begin{subfigure}[t]{0.3\linewidth}
    \centering
    \includegraphics[width=\linewidth]{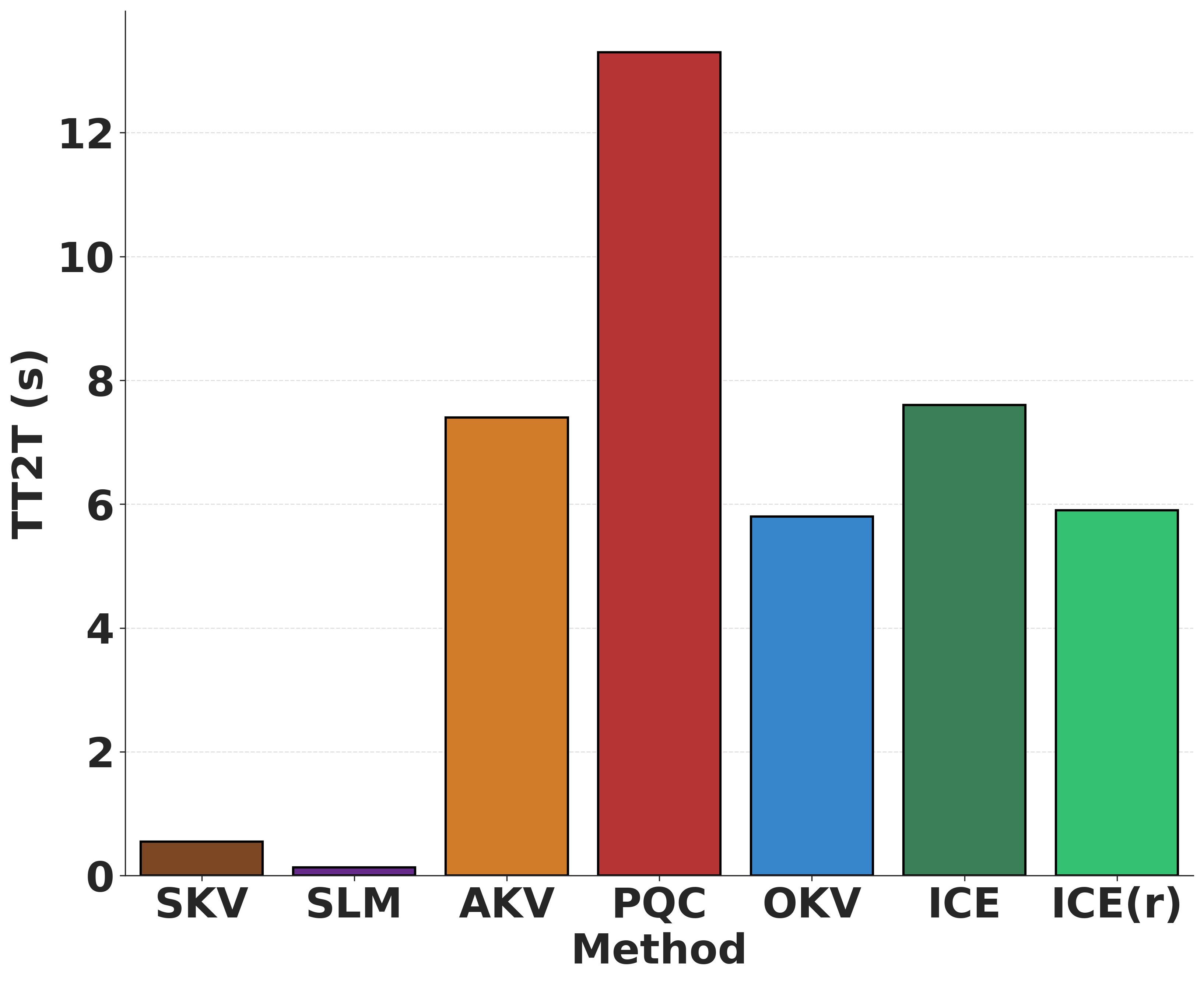}
    \caption{Time to the second token.}
    \label{fig:tt2t}
\end{subfigure}
\hfill
\begin{subfigure}[t]{0.3\linewidth}
    \centering
    \includegraphics[width=\linewidth]{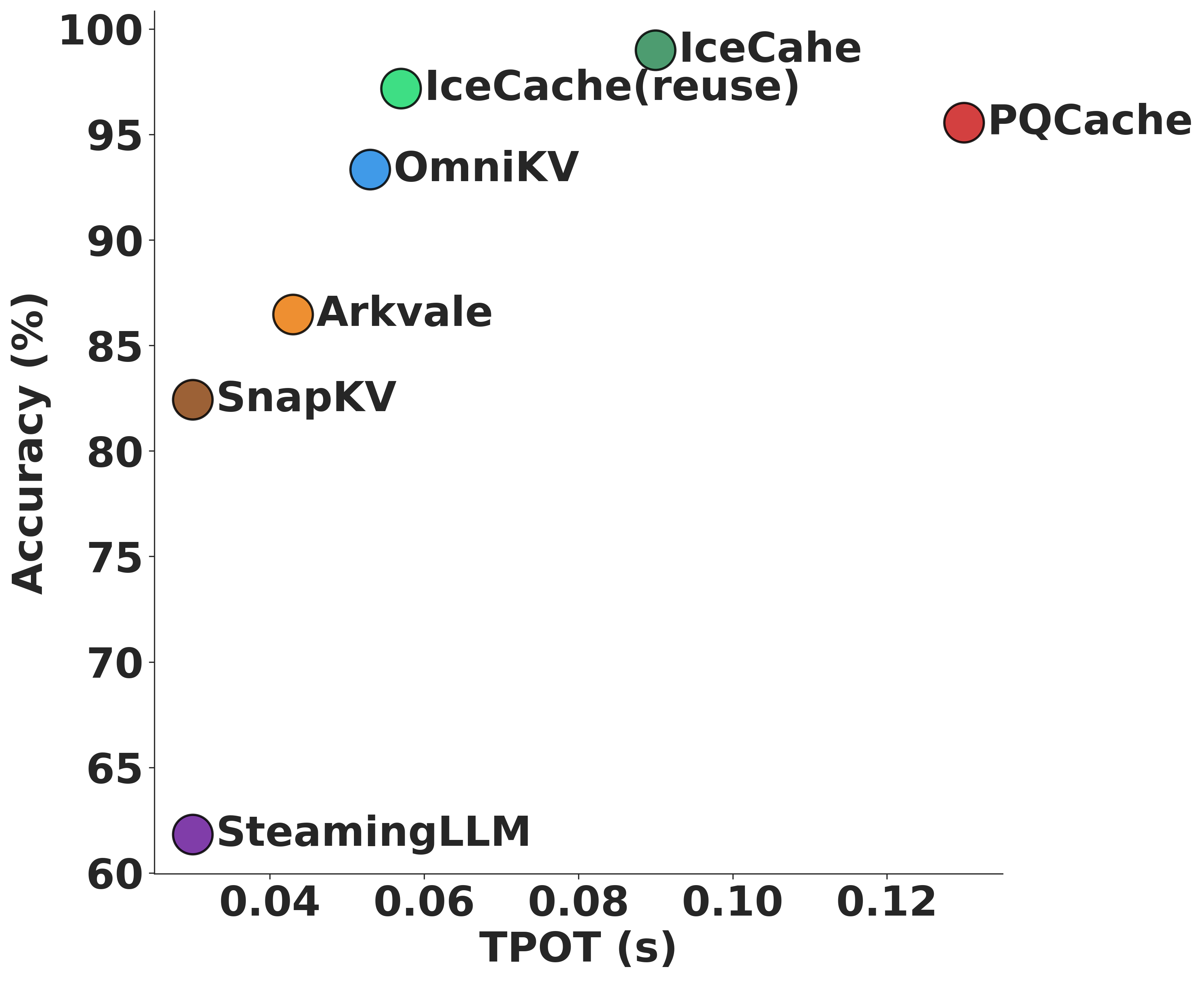}
    \caption{Time per output token.}
    \label{fig:throughput}
\end{subfigure}
\hfill
\begin{subfigure}[t]{0.3\linewidth}
    \centering
    \includegraphics[width=\linewidth]{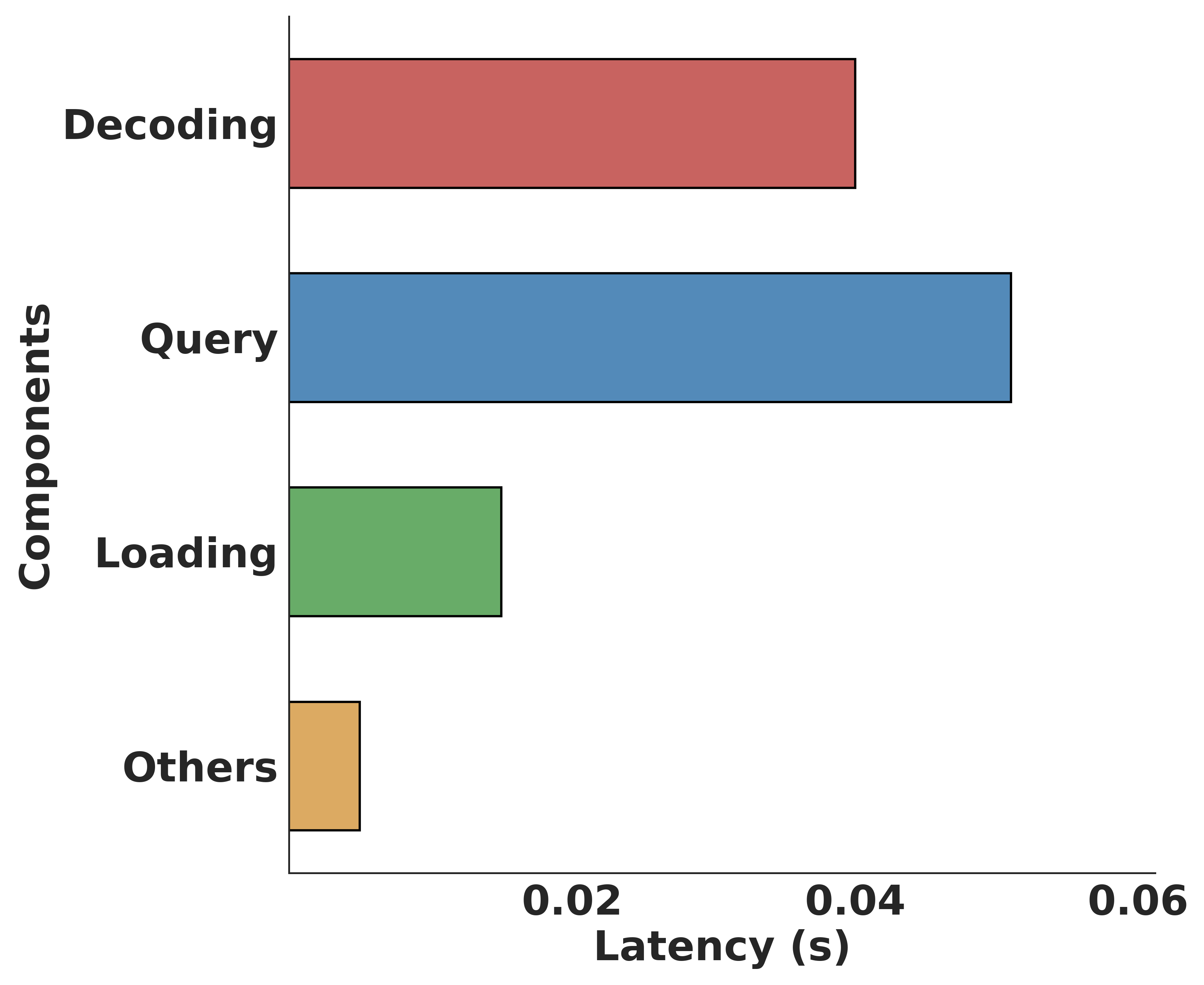}
    \caption{TPOT Latency breakdown.}
    \label{fig:breakdown}
\end{subfigure}
\caption{
Latency comparison of \toolName and baseline methods on a 36k-token sequence.
}
\label{fig:latency}
\end{figure}

\vspace{-1mm}
Since most baselines, including \toolName, use the entire KV-cache to generate the first token, we follow PQCache~\citep{zhang2024pqcache} and report the Time to the second token (TT2T) in Fig.~\ref{fig:tt2t}, for a 36k sequence length with Llama-3.1-8B-Instruct and all the methods compared in Table~\ref{table:longbench}, except MagicPig, which is restricted to
costly AVX-512 CPUs.
Additionally, we present results for \toolName incorporating the critical-page-reuse technique (reusing the same selected KV-page indices across layers), a method also used in OmniKV, which serves as an approximate version that trades accuracy for increased speed. 

As illustrated in the figure, our method, \toolName, achieves competitive latency among retrieval-based algorithms, recording 7.7 seconds. \toolName(reuse) further reduces this to 5.9 seconds, matching OmniKV (5.8 s) and outperforming other retrieval-based baselines such as Arkvale (7.4 s) and PQCache (13.3 s). Although eviction-based methods like SnapKV (0.55 s) and StreamingLLM (0.13 s) are much faster, their speed often comes at the cost of accuracy. Overall, IceCache and IceCache(reuse) offer a strong balance between efficiency and accuracy, with the reuse variant showing how our approach can further optimize latency without significantly sacrificing performance. We include more details of IceCache(reuse) in the appendix~\ref{sec:reuse}.

Similarly, for decoding latency, Fig.~\ref{fig:throughput} shows the average time per generated token, along with the corresponding accuracy percentage relative to the full KV-cache model, for an input sequence length of 36k. Eviction-based methods, StreamingLLM and SnapKV (both at 0.03 seconds per token), continue to show the fastest speeds due to their minimal overhead. Among the more accurate retrieval-based methods, IceCache(reuse) achieves a highly competitive decoding time of 0.06 seconds per token -- substantially faster than PQCache (0.13 s) and nearly matching the speed of OmniKV (0.05 s). Vanilla \toolName maintains a strong balance, combining superior accuracy with efficient decoding. It achieves the highest accuracy percentage (99.0\%) while still outperforming PQCache in speed. These results further demonstrate that \toolName effectively balances accuracy and decoding latency.

Figure~\ref{fig:breakdown} presents a detailed breakdown of TPOT latency for \toolName at a sequence length of 36k, with a total latency of 0.11 seconds. In this figure, “Loading”, “Query”, and “Decoding” correspond to the overhead from CPU–GPU communication, DCI-query operations, and the overall LLM decoding process, respectively. The largest contributors to latency are the DCI-query module (0.05 s) and decoding (0.04 s), while GPU–CPU offloading and other miscellaneous operations add only 0.015 seconds and 0.005 seconds, respectively.

\vspace{-2mm}
\subsection{GSM8k CoT Reasoning}
\vspace{-1mm}
\begin{wraptable}{r}{0.3\textwidth}
\vspace{-6mm}
    \centering
    \caption{GSM8K CoT.}
    \scalebox{0.612}{
    \begin{tabular}{lcc}
    \toprule
    Method & Budget & Accuracy \\
    \midrule
    FULL & N/A & 48.2 \\
    \midrule
    SKV & $10\%$ & 44.7 \\
    SLM & $10\%$ & 44.4 \\
    OKV & $10\%$ & 42.7 \\
    MPG & $10\%$ & 43.1\\
    PQC & $10\%$ & 46.2\\
    AKV & $10\%$ & 30.9\\
    \midrule
    \rowcolor{green!10}
    ICE & $10\%$ & \textbf{47.4}\\
    \bottomrule
    \end{tabular}
    }
\label{tab:gsm}
\end{wraptable}
We also evaluate \toolName on the GSM8K benchmark using Chain-of-Thought prompting, applying a 10\% budget for all compared methods using Mistral-7B-Instruct-v0.2.
As shown in Table~\ref{tab:gsm}, \toolName demonstrates superior performance, achieving an accuracy of 47.4\%, significantly higher than all other methods under the same budget constraint. In particular, it improves upon the strongest baseline, PQCache (46\%), by 1.2\% absolute points, reaching 47.4\%.  Moreover, our approach nearly matches the full KV-cache (48.2\%), highlighting the effectiveness of \toolName.
\vspace{-2mm}
\subsection{Scalability to Extremely Long Context}
\begin{figure}[!ht]
\centering
\begin{minipage}[t]{0.42\linewidth}
\centering
\vspace{3mm}
\includegraphics[width=\linewidth]{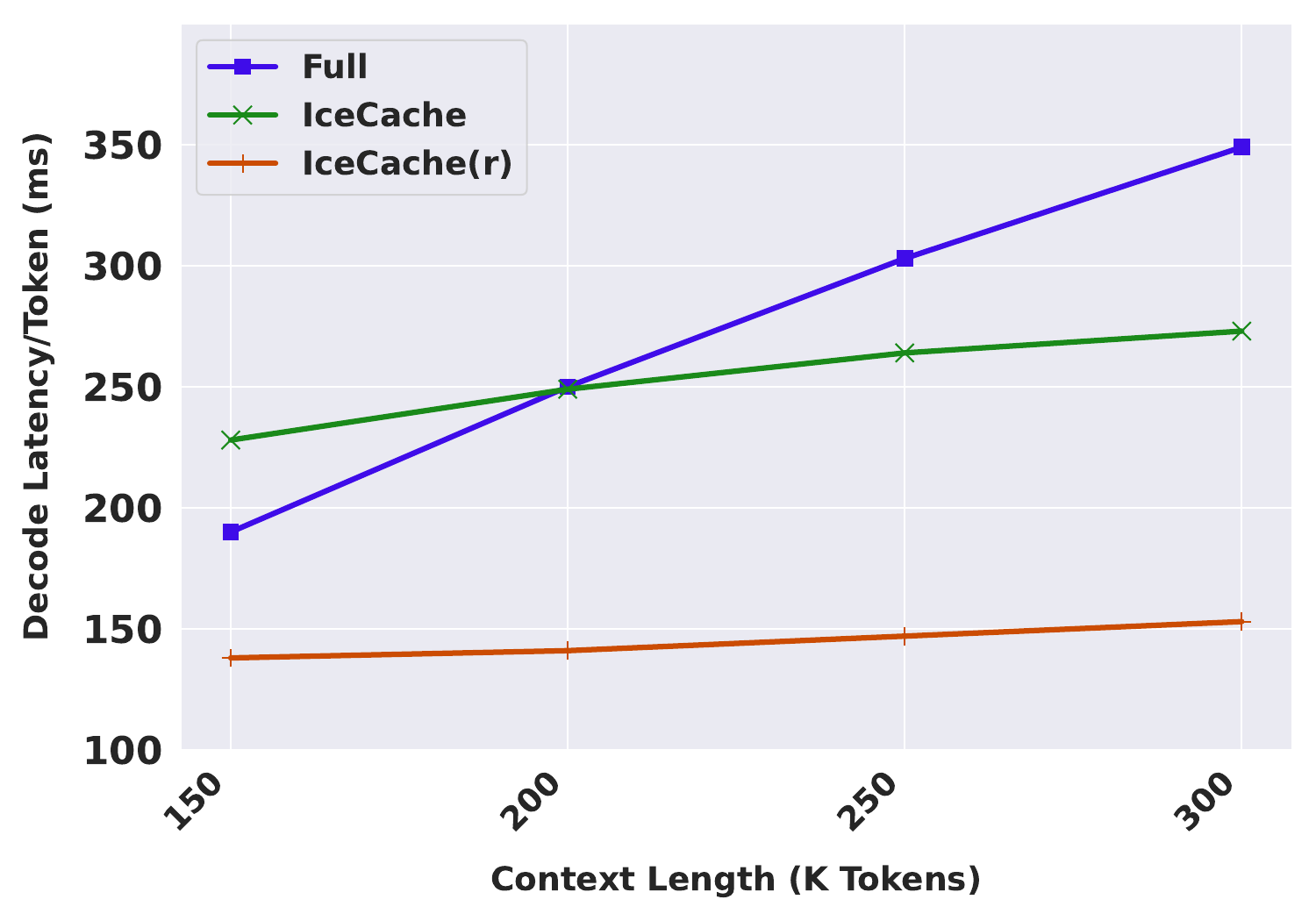}
\caption{Latency scaling across context lengths (150k, 200k, 250k and 300k).}
\label{fig:scaling}
\end{minipage}
\hfill
\begin{minipage}[t]{0.42\linewidth}
\centering
\vspace{-2mm}
\captionof{table}{Accuracy comparison on Qwen3-4B-Instruct under different context lengths (150k, 200k, 250k) on RULER benchmark.}
\label{tab:icecache_qwen3}
\scalebox{0.6}{
\begin{tabular}{lccc}
\toprule
\textbf{Method} & \textbf{Single-NIAH} & \textbf{Multi-keys NIAH} & \textbf{QA} \\
\midrule
\multicolumn{4}{c}{\textbf{150k Tokens}} \\
\midrule
Full KV        & 100.0 & 97.0 & 93.0 \\
IceCache       & 100.0 & 98.0 & 92.0 \\
IceCache (r)   & 100.0 & 98.0 & 91.0 \\
\midrule
\multicolumn{4}{c}{\textbf{200k Tokens}} \\
\midrule
Full KV        & 100.0 & 92.0 & 95.0 \\
IceCache       & 100.0 & 91.3 & 95.0 \\
IceCache (r)   & 100.0 & 91.0 & 94.3 \\
\midrule
\multicolumn{4}{c}{\textbf{250k Tokens}} \\
\midrule
Full KV        & 100.0 & 91.0 & 91.3 \\
IceCache       & 100.0 & 93.0 & 91.7 \\
IceCache (r)   & 100.0 & 92.0 & 92.0 \\
\bottomrule
\end{tabular}
}
\end{minipage}
\end{figure}

\vspace{-1mm}
To further evaluate the performance of \toolName under extremely long-context settings, we conduct experiments on the RULER benchmark and report accuracy on Single-Needle-in-a-Haystack (Single-NIAH), Multi-keys NIAH, and QA tasks with context lengths of 150k, 200k, and 250k tokens, using a token budget of 256. The experiments are conducted using Qwen3-4B-Instruct-2507 on a single H100 GPU with 64 CPU threads enabled. As shown in Table~\ref{tab:icecache_qwen3}, \toolName and IceCache(r) consistently maintain accuracy comparable to Full-KV across all tasks and context lengths. Figure~\ref{fig:scaling} further demonstrates that, as the input context grows, both \toolName and IceCache(r) exhibit a substantially slower increase in per-token decoding latency than Full-KV, whose decoding cost scales sharply with sequence length.
These results indicate that \toolName achieves a better accuracy–latency trade-off for extremely long-context inference, enabling scalable decoding without sacrificing accuracy on the RULER benchmark.

\vspace{-5mm}
\section{Conclusion}
\vspace{-3mm}
This paper addresses the fundamental challenge of long-context inference in LLMs, where the rapidly growing KV-cache consumes substantial GPU memory and degrades computational efficiency. We first introduce the DCI-tree, a hierarchical indexing structure that integrates naturally with PagedAttention and supports dynamic updates for efficient KV-cache organization. Building on this foundation, we propose \toolName, a page-based KV-cache management framework that combines semantic token clustering with efficient GPU–CPU offloading. Across diverse long-context benchmarks, \toolName consistently achieves superior accuracy–latency trade-offs. These results establish \toolName as a scalable and practical solution for memory-efficient long-context LLM inference.

\section{Acknowledgements}
We thank the anonymous reviewers for their insightful feedback and constructive suggestions. This research was enabled in part by support provided by NSERC, the BC DRI Group and the Digital Research Alliance of Canada.


\bibliography{iclr2026_conference}
\bibliographystyle{iclr2026_conference}

\newpage
\appendix
\section{Method Overview}\label{sec:code}
We separate all tokens into three groups: \emph{sink tokens}, which are the tokens at the very beginning of the input sequence; \emph{window tokens}, which are the most recent tokens; and all the remaining tokens in between. The pages that store sink tokens are referred to as \emph{sink pages}, and those that store window tokens are referred to as \emph{window pages}. We always keep all the sink and window pages in GPU. 

We provide the pseudocode below for \toolName. It operates in two main phases: (1) Prefill Phase: During the initial processing of prompt tokens, \toolName allocates paged KV memory per layer and performs self-attention computations. From the third layer onward, it copies KV embeddings to CPU and builds a dynamic index -- DCI-tree. This tree enables efficient future lookup of important tokens based on query embeddings. (2) Decode Phase: During autoregressive decoding, each new token’s query embedding is used to retrieve the most relevant KV pages via DCI-based query. Selected pages are back-loaded to GPU on demand, while unimportant pages are offloaded to CPU storage. When a new window page is offloaded, the DCI-tree is incrementally updated to store tokens in this page. The detailed steps are outlined in Algorithm~\ref{alg:kv-cache}. We will explain the mechanisms behind indexing and page selection in the following sections.

\section{Details of IceCache (reuse) on LongBench}\label{sec:reuse}
We present the LongBench scores for \toolName (reuse) in Table~\ref{table:reuse}. Starting from the third layer, we build the DCI-tree and perform DCI-queries every three layers; we refer to these as “anchor layers”. For the layers in between, we reuse the KV-cache indices selected at the most recent anchor layer.
\begin{table*}[h!]
\centering
\caption{Accuracy of IceCache (reuse) on LongBench.}
\footnotesize  
\setlength{\tabcolsep}{2pt}  
\scalebox{0.612}{
\begin{tabular}{c|c|c|c|c|c|c|c|c|c|c|c|c|c|c|c|c|c|c}
\midrule
\multirow{3}{*}{\textbf{Budget}} & \multirow{3}{*}{\textbf{Method}} & \multicolumn{3}{c|}{\textbf{Single-Document QA}} & \multicolumn{3}{c|}{\textbf{Multi-Document QA}} & \multicolumn{3}{c|}{\textbf{Summarization}} & \multicolumn{3}{c|}{\textbf{Few-shot Learning}} & \multicolumn{2}{c|}{\textbf{Synthetic}} & \multicolumn{2}{c|}{\textbf{Code}} & \multirow{3}{*}{\textbf{Avg.}} \\
\cmidrule{3-18}
& & NrtvQA & Qasper & MF-en & HotpotQA & 2WikiMQA & Musique & GovReport & QMSum & MultiNews & TREC & TriviaQA & SAMSum & PCount & PRe & Lcc & RB-P & \\
& & 18409 & 3619 & 4559 & 9151 & 4887 & 11214 & 8734 & 10614 & 2113 & 5177 & 8209 & 6258 & 11141 & 9289 & 1235 & 4206 & \\
\midrule
\midrule
\multicolumn{19}{c}{\textbf{Llama-3.1-8B-Instruct}} \\
\midrule
256 & ICE(r) & 29.1 & 43.6 & 53.4 & 55.1 & 44.3 & 29.0 & 31.8 & 24.3 & 26.2 & 72.0 & 91.1 & 41.0 & 7.2 & 100.0 & 60.7 & 53.1 & 47.7 \\
\midrule
\end{tabular}
}
\label{table:reuse}
\end{table*}
\section{Performance on Long Generation Benchmark}
To evaluate the effectiveness of \toolName on long-context generation tasks, we conduct experiments on LongGenBench~\citep{wu2024longgenbench} using Llama-3.1-8B-Instruct with a 256-token budget. As shown in Table~\ref{tab:longgen}, \toolName substantially outperforms PQCache while maintaining accuracies on par with the Full-KV baseline.
\begin{table}[!h]
\centering
\caption{Accuracy comparison on LongGenBench for Llama-3.1-8B-Instruct.}
\tiny
\begin{tabular}{lccccc}
\toprule
Method & Completion Rate & Accuracy Once & Accuracy Range & Accuracy Periodic & Avg. Accuracy \\
\midrule
Full KV & 97.627 & 0.349 & 0.488 & 0.135 & 0.324 \\
\toolName & 95.322 & 0.377 & 0.454 & 0.164 & 0.331 \\
PQCache & 88.691 & 0.318 & 0.395 & 0.105 & 0.273 \\
\bottomrule
\end{tabular}
\label{tab:longgen}
\end{table}

\section{LLM Usage}
This work focuses on optimizing KV-cache management for large language models (LLMs). All the base models used in this paper can be viewed as LLMs, including Llama-3.1-8B-Instruct, Mistral-7B-Instruct-v0.2, Qwen3-32B, LongChat-7B-v1.5 and Qwen3-4B-Instruct-2507. We also leveraged an LLM to help refine the manuscript’s language and improve its overall readability.

\newpage
\begin{algorithm}[H]
\caption{\toolName}
\label{alg:kv-cache}
\begin{algorithmic}[1]
\State \textbf{Input:} Sequence of tokens $x_{1:I}$, Transformer with $L$ attention layers, Page size $s$
\State \textbf{Phase 1: Prefill}
\For{$\ell = 0$ to $L-1$}
    \State Allocate pages and arrange KVs to the pages for layer $\ell$
    \If{$\ell \geq 2$}
        \State Copy KVs of tokens between sink tokens and window tokens from GPU to CPU (denoted as $S_k$ and $S_v$)
    \EndIf
    \State Compute the output from the current self-attention layer $\ell$
    \If{$\ell \geq 2$}
        \State $T_l \gets$ DCI-INDEXING($S_k,S_v$)
    \EndIf
\EndFor

\State \textbf{Phase 2: Decode (repeated over time steps $i > I$)}
\While{receive new token $x_i$ with $\mathbf{q}_i$ as its query embedding}
    \For{$\ell = 0$ to $L-1$}
        \If{Number of tokens in the last page $\geq s - 1$ }
            \State Offload the oldest window page $P_w$ from GPU to CPU
            \State Set Flag to True
        \EndIf
        \State Append KVs of $x_i$ to the end of the newest window page
        \If{$\ell \geq 2$}
            \State $S_{l} \gets$ PAGE-SELECT($\mathbf{q}_i, T_l, k$)
            \State Recall selected pages $S_{l}$ from CPU to GPU
        \EndIf
        \State Compute the output from the current self-attention layer $\ell$
        \If{$\ell \geq 2$ \& Flag is True}
        {
            \State {// Insert the tokens in offloaded $P_w$ to $T_l$}
            \For{$i$ \textbf{in} $P_w$}
                \State {{// Random Promotion}}
                \State $l_i \gets 1$ \Comment{Start at bottom level}
                \While{$\text{Random}(0, 1) < r$}
                    \State $l_i \gets l_i + 1$ \Comment{Promote to the next higher level}
                \EndWhile
                \State $\{p_{i}\} \gets $ QUERY$(\mathbf{k}_i,T_l,l_i,1)$ \Comment{Get the parent index using QUERY (Alg.4)}
                \State Insert $\mathbf{k}_i$ to $T_l$ with $p_{i}$ as its parent node index
            \EndFor
            }
        \EndIf
    \EndFor
    \If{$x_i$ = EOS}
        \State Break
    \EndIf
    \State $i=i+1$
\EndWhile
\end{algorithmic}
\end{algorithm}
\begin{algorithm}[H]
\footnotesize
\caption{Indexing}
\label{alg_prompt}
\begin{algorithmic}
\Require A list $S_k$ of $n$ keys $\mathbf{k}_1,\ldots,\mathbf{k}_n \in \mathbb{R}^{d}$, Promotion ratio $r$
\Function{DCI-Indexing}{$S_k, r$}
    {
    \For{$i = 1$ \textbf{to} $n$}
        \State {// Random Promotion}
        \State $l_i \gets 1$ \Comment{Start at bottom level}
        \While{$\text{Random}(0, 1) < r$}
            \State $l_i \gets l_i + 1$ \Comment{Promote to the next higher level}
        \EndWhile
    \EndFor}
    \State {Remove the empty levels}
    \State Initialize $T$ with an empty root node
    \For{$i = 1$ \textbf{to} $n$}
        \State $\{p_{i}\} \gets $ QUERY$(\mathbf{k}_i,T,l_i,1)$ \Comment{{Get the parent index using QUERY (Alg.4)}}
        \State Insert $\mathbf{k}_i$ to $T$ with $p_{i}$ as its parent node index
    \EndFor
    \State \Return $T$
\EndFunction
\end{algorithmic}
\normalsize
\end{algorithm}
\begin{algorithm}[H]
\footnotesize
\caption{Page Selection}
\label{alg_page_select}
\begin{algorithmic}
\Require Query vector $\mathbf{q}_i \in \mathbb{R}^{d}$, DCI-tree $T$, Number of critical keys $k$
\Function{PAGE-SELECT}{$\mathbf{q}_i, T, k$}
    \State Initialize $S_{l} \gets \emptyset$
    \State $S_{k} \gets$ QUERY$(\mathbf{q}_i, T, -1, k)$
    \State $S_{l} \gets$ \text{FIND-PAGE-INDEX}$(S_{k})$
    \State \Return $S_{l}$
\EndFunction
\end{algorithmic}
\normalsize
\end{algorithm}
\begin{algorithm}[H]
\footnotesize
\caption{$k$-Nearest Neighbour Querying}
\label{alg_query}
\begin{algorithmic}
\Require Query vector $\mathbf{q}_i \in \mathbb{R}^{d}$, DCI-tree $T$ with L levels, Target level $l$, Number of critical keys $k$
\Function{Query}{$\mathbf{q}_i, T, l, k$}
    \If{$l = -1$}
        \State $l \gets 1$
        \State Set Flag to True
    \Else
        \State Set Flag to False
    \EndIf
    \State $S \gets \emptyset$
    \State $P \gets$ empty priority queue with size $k$
    \For{$i = L$ \textbf{to} $l$}
        \State $S' \gets \emptyset$
        \If{$i = L$}
            \State $S \gets \{$root node$\}$
        \EndIf
        \For{$s$ \textbf{in} $S$}
            \State $S'' \gets$ Prioritized-DCI-Query$(\mathbf{q}_i, s, k)$
            \State $S' \gets S' \cup S''$
        \EndFor 
        \If{Flag is True \textbf{or} $i = l$}
            \For{$s$ \textbf{in} $S'$}
                \State $P \gets$ Add-to-Priority-Queue $(P, s)$
            \EndFor 
        \EndIf
        \State $S \gets S'$
    \EndFor
    \State \Return $k$ nodes in $P$ that have the keys with maximum inner-product with $\mathbf{q}_i$
\EndFunction
\end{algorithmic}
\normalsize
\end{algorithm}

\end{document}